\documentclass[sigconf]{acmart}
\usepackage{hyperref}
\usepackage{graphicx}
\usepackage{subcaption}

\AtBeginDocument{%
  }

\copyrightyear{2026}
\acmYear{2026}
\setcopyright{cc}
\setcctype{by}

\acmConference[NordiCHI '26 Adjunct]{Adjunct Proceedings of the 14th Nordic Conference on Human-Computer Interaction}{October 03--07, 2026}{Vaasa, Finland}

\acmBooktitle{Adjunct Proceedings of the 14th Nordic Conference on Human-Computer Interaction (NordiCHI '26 Adjunct), October 03--07, 2026, Vaasa, Finland}

\acmDOI{10.1145/3821402.3830156}
\acmISBN{979-8-4007-2798-6/2026/10}

\begin{document}

%%
%% The "title" command has an optional parameter,
%% allowing the author to define a "short title" to be used in page headers.
\title{Man, Machine, and Masterpiece: Artistic Ownership in the AI Era}

%%
%% The "author" command and its associated commands are used to define
%% the authors and their affiliations.
%% Of note is the shared affiliation of the first two authors, and the
%% "authornote" and "authornotemark" commands
%% used to denote shared contribution to the research.
\author{Sofi Gjing Jovanovska}
\email{Sofi.Jovanovska@student.uni-siegen.de}
\affiliation{%
  \institution{University of Siegen}
  \streetaddress{Kohlbettstraße 15}
  \city{Siegen}
  \postcode{57072}
  \country{Germany}
}

\author{Kuntal Ghosh}
\email{kuntal.ghosh@uni-siegen.de}
\affiliation{%
  \institution{University of Siegen}
  \streetaddress{Kohlbettstraße 15}
  \city{Siegen}
  \postcode{57072}
  \country{Germany}
}

\author{Daniel Muhu Njenga}
\email{Daniel.Njenga@student.uni-siegen.de}
\affiliation{%
  \institution{University of Siegen}
  \streetaddress{Kohlbettstraße 15}
  \city{Siegen}
  \postcode{57072}
  \country{Germany}
}

\author{Ahmed Mufassir}
\email{Ahmed.MohamedAmeen@student.uni-siegen.de}
\affiliation{%
  \institution{University of Siegen}
  \streetaddress{Kohlbettstraße 15}
  \city{Siegen}
  \postcode{57072}
  \country{Germany}
}

\author{Shadan Sadeghian}
\email{shadan.sadeghian@uni-siegen.de}
\affiliation{%
  \institution{University of Siegen}
  \streetaddress{Kohlbettstraße 15}
  \city{Siegen}
  \postcode{57072}
  \country{Germany}
}

%%
%% By default, the full list of authors will be used in the page
%% headers. Often, this list is too long, and will overlap
%% other information printed in the page headers. This command allows
%% the author to define a more concise list
%% of authors' names for this purpose.
\renewcommand{\shortauthors}{Jovanovska et al.}

%%
%% The abstract is a short summary of the work to be presented in the
%% article.
\begin{abstract}
    The integration of AI-driven systems in creative work has sparked debates among artists and legal communities about notions of ownership.
    Yet there remains little consensus on how ownership should be defined and attributed when human and AI contributions are intertwined.
    To provoke critical reflection on these tensions, we designed ArtSplit, a provotype that explicitly quantifies human and AI contributions across different stages of creative work. Rather than aiming to resolve ownership, the provotype was used to elicit artists’ responses to the idea of attributing ownership through measurable actions in the creative workflow.
    We argue that quantification fails to align with artists’ understandings of creative intent and agency, and that efforts to measure ownership risk diluting long-standing assumptions through which artists understand and practice creative work. This critique challenges the impulse to transform a historically and socially situated relation into a technical problem.
\end{abstract}

%%
%% The code below is generated by the tool at http://dl.acm.org/ccs.cfm.
%% Please copy and paste the code instead of the example below.
%%

%%
%% Keywords. The author(s) should pick words that accurately describe
%% the work being presented. Separate the keywords with commas.

% \received{20 February 2007}
% \received[revised]{12 March 2009}
% \received[accepted]{5 June 2009}

%%
%% This command processes the author and affiliation and title
%% information and builds the first part of the formatted document.
\maketitle

\section{Introduction} 
As early as 2018, artists such as Robbie Barrat have been experimenting with AI art\footnote{\url{https://stanforddaily.com/2018/06/12/qa-robbie-barrat-on-training-neural-networks-to-create-art/}}. Yet, around 2022, the use of AI in art began sparking intense controversies. Two things happened that year: AI image-generators such as Stable Diffusion\footnote{\url{https://stability.ai/}}, Mid-journey\footnote{\url{https://www.midjourney.com/home}}, and Dall-E\footnote{\url{https://openai.com/index/dall-e/}} were released for public use, which made AI in artistic work too prevalent to ignore. And second, Jason M. Allen won a prestigious art award using an AI-generated image\footnote{\url{https://edition.cnn.com/2022/09/03/tech/ai-art-fair-winner-controversy}}. Artists have expressed anger about the possibility of their work being used to train these systems\footnote{\url{https://www.theguardian.com/artanddesign/2023/jan/23/its-the-opposite-of-art-why-illustrators-are-furious-about-ai}}. Many recalled Studio Ghibli’s founder saying after seeing an AI animation that "it feels like an insult to life itself"\footnote{\url{https://faroutmagazine.co.uk/hayao-miyazaki-on-ai-utterly-disgusted/}}. While the critiques of AI art come from many angles, one of the most commonly raised questions is: \textit{Can the artist who generated their artwork with AI be considered the owner of the work?}

Automating parts or entire processes of work is nothing new. Moreover, artistic work, as a form of labor, shares many basic characteristics with 'ordinary' work: the worker, here the artist, produces, usually through substantial effort, something that others can consume, in this case the audience; when the product is deemed capable of satisfying some desires of consumers, the artist is paid, either directly by institutions acting as employers or by the commissioner who is also the consumer, or in cases where the artist’s capacities are highly recognized, through an advance payment. If producing something consumers can enjoy is all that matters, then using AI in artistic work should not cause special issues — at least, no different from those raised by using AI in other forms of labor.
However, many are unwilling to grant that artistic work permits any form of automation. The reason is that artistic work is often perceived as different from other kinds of work, as the artist bears a unique and personal relation to the artwork. 
When an ordered cake is delivered, the baker is no longer tied to it (except in cases like contamination or if the baker is a celebrity). We also do not usually care if automation was involved. By contrast, artistic work continues to credit the artist even after it reaches the audience. 
The artist typically signs the work or, at any rate, is credited for the work somewhere as its creator. This relation is called authorship: a special way of being the owner of the artwork \cite{irvin2005appropriation}. The artist is the author of their artwork even when they own neither copyright nor the physical object or digital token, because the artwork has resulted from their creativity. And it is because of this idea that the use of AI in artistic work is controversial -- \textit{Can the artist still be considered the author if an AI system helped generate the image?}

Two common reasons opponents of AI art are unwilling to see artists as authors are: first, it is the machine, instead of the human artist, that does the “generating”; and second, the data provided by other artists is often used to feed the machine \cite{2023AIImpact}. For the first reason, one could say that the artist certainly has done something during the creative process - at the very least, has created prompts and made iterative refinements \cite{2024Intenttuner}. The real issue here, one may suspect, is that the work done by the machine is not transparent enough for us to assess whether it is of more worth compared to the work done by the artist \cite{xu2024makes}. For the second issue, one could still question whether a typical good case of artistic creation involves no input from others. The overarching question here is in AI-mediated creative work, what kinds of contributions are understood to warrant ownership, and whether there exists a threshold at which an artist’s involvement is considered sufficient to merit authorship or credit.

In more traditional models of creative work, the relation between artist and artwork has been articulated and stabilized through legal frameworks such as copyright, which presume that ownership originates in a human creator whose creative decisions can be traced to the final work. Here (and later in this critique), we use the term \textit{ownership} to refer to an intellectual or symbolic claim through which an artist is recognized as the creator of a work, encompassing what is often described as authorship, such as the attribution of creative origin, responsibility, and identity, rather than legal or material ownership.

With the growing use of generative AI in artistic practice, however, this presumption has become increasingly difficult to maintain. Legal and regulatory bodies (most notably in the context of the EU-AI Act \cite{eu_ai_act_2024}) have begun revisiting how ownership should be understood when creative work is produced through a combination of human effort and automated generation. What these discussions reveal is not a clear alternative model, but rather the limits of existing ones: copyright regimes remain largely oriented toward singular human ownership and struggle to account for creative processes in which agency is distributed across people, tools, data, and algorithms \cite{guadamuz2017ai, irvin2005appropriation}.
In response, regulatory discourse often falls back on notions such as \textit{“significant human contribution”}, suggesting that ownership depends on whether a human has exercised sufficient creative control or left a recognizable personal imprint on the work \cite{xiao2023decoding, ramos2025reconceptualizing}. Yet these criteria remain qualitative and loosely specified. They offer little insight into how different kinds of involvement—conceptual framing, prompt crafting, selection, iteration, or post-processing—should be compared, weighted, or even recognized in practice. This ambiguity becomes especially salient in AI-mediated creative work, where human and machine contributions are interwoven over time and cannot easily be separated.

Rather than treating these ambiguities as problems to be resolved, this critique paper approaches them as sites of inquiry. We intend to understand how artists make sense of ownership when working with AI, and how attempts to render ownership visible, particularly through quantitative or system-generated representations, shape their perceptions of creative ownership. To this end, we designed an AI-driven provotype \cite{boer2012provotypes} that deliberately operationalizes ownership in different ways, not as definitive solutions, but as provocations. Through interviews supported by video vignettes of these provotypes, we explore how such framings affirm, challenge, or destabilize artists’ sense of ownership.

\section{Background} \label{sec:background}

Creative work has long been understood as producing new and meaningful expressions across domains such as visual art \cite{gerber2020work}, literature \cite{doyle1998writer}, music \cite{cook2018music}, design \cite{bowen2016value}, and more \cite{brace2010recovering, hesmondhalgh2008creative, goldsmith2018embedded}. Beyond originality and aesthetic value \cite{boden2004creative, runco2012standard}, creative practice is shaped by broader contextual, cultural, and collaborative influences \cite{oldham1996employee, erez2010creativity, abra1994collaboration}, and is closely tied to how individuals understand themselves and their place in the world. Artists describe activities such as painting, writing, composing, or performing as reflective practices through which they explore and express their sense of self \cite{lingo2013looking}. Through creative work, artists articulate values \cite{taylor2016contemporary}, emotions \cite{radford2004emotion}, and worldviews \cite{wilson2010social}, shaping narratives of who they are and what they stand for \cite{beech2016identity}.

This relationship between creativity and identity is often entangled with questions of \textit{"ownership"} and \textit{"authorship"}. Although these terms are frequently used interchangeably, they carry distinct legal and conceptual meanings \cite{xu2024makes}.
Ownership extends beyond legal recognition to encompass a subjective sense of personal attachment (psychological ownership), through which creators experience an idea or artifact as theirs \cite{wu2023owndiffusion, pierce2003state}. This sense of ownership is frequently shaped through involvement in the creative process and the act of bringing something into existence, whether tangible or intangible \cite{baer2012blind}.
Authorship, however, remains a more fluid and contested notion in artistic practice, where formalized rules are often absent, and claims to authorship are frequently asserted by artists themselves rather than determined by external criteria \cite{cronin2012collaboration}.
Research in cultural and social psychology challenges individualistic notions of creativity by emphasizing the role of social interaction, communication, and collaboration in creative work \cite{barrett2021creative, sawyer2009distributed}. Creative outcomes frequently emerge through group processes, shared cognition, and mutual inspiration rather than isolated individual effort \cite{paulus2000groups}. In such contexts, the boundaries between ownership and authorship often blur as ideas are jointly developed through co-creation and iterative input \cite{dreyfuss2000collaborative}. Yet, despite these collaborative realities, enduring cultural narratives—such as the figure of the lone artistic genius—continue to shape expectations around creative credit and recognition \cite{cronin2012collaboration}.

%Introducing AI:-
As AI-driven systems become increasingly embedded in creative practices, questions of creativity, ownership, and contribution have become more difficult to settle \cite{anantrasirichai2022artificial}. Artists often describe AI-generated outputs as surprising in ways that sometimes exceed direct human intention, which human creators curate, refine, and reinterpret \cite{shelby2024generative}. Such interactions redistribute creative initiative across human and non-human actors, unsettling familiar assumptions about artistic agency and control. Moreover, AI-assisted generation has been shown to reduce the burden of labor-intensive creative tasks and support coordination and shared understanding in collaborative work \cite{han2024teams}. Yet these affordances have not led to consensus about how AI-assisted art should be understood or valued.

One common position holds that AI should be treated as a tool like a camera or other creative medium, through which artists pursue their intentions, rather than as an author in its own right \cite{hertzmann2018can, hertzmann2020computers}. From this view, attributing authorship to AI risks obscuring the human labor involved and diminishing the artist’s role \cite{grba2021brittle, grba2022deep}. Others point out that contemporary AI-based practices involve substantial creative work such as prompt formulation, iterative experimentation, and curatorial decision-making, all of which shape the outcome in meaningful ways \cite{chang2023prompt}. Still, empirical studies reveal persistent concerns among creators and audiences that AI-generated works lack intention, emotional investment, and experiential qualities traditionally associated with human creativity \cite{shelby2024generative, guo2024exploringevolvement, hwang2022too}. 

Underlying these disagreements is a persistent question of attribution: To what extent can a creative outcome be credited to the human artist rather than the AI-driven system \cite{samuelson1985allocating, hertzmann2020computers, eshraghian2020human, wu2023owndiffusion}? As AI-generated works are frequently framed as products of complex algorithms rather than individual minds \cite{oksanen2023artificial}, creators often struggle to draw clear boundaries \cite{chang2023prompt}. Rather than appealing to measurable contributions, they tend to reason about ownership in terms of conceptual framing, content gatekeeping, and final decision-making authority \cite{chang2023prompt, hwang202580}. Psychological ownership in these contexts is often shaped by factors such as perceived involvement, expectations of credit, and concerns about infringement \cite{xu2024makes}. Concurrently, legal frameworks offer little concrete guidance on human–AI collaboration \cite{zirpoli2023generative, watiktinnakorn2023blurring}, leaving questions of authorship to be negotiated through personal judgment rather than established rules.

From a legal perspective, these ambiguities are not easily resolved. In the European Union, copyright protection hinges on whether a work is original in the sense that it reflects the author’s own intellectual creation \cite{eprs2025ai}. Case law from the Court of Justice of the European Union emphasizes that originality arises when an author makes free and creative choices that are visible in the final form of the work \cite{ramos2025reconceptualizing}. Copyright protection, therefore, does not depend on effort or contribution alone, but on whether human creative judgment can be identified in the expression of the work \cite{bently2022intellectual}. However, EU-law offers no explicit definition of authorship, leaving such determinations to interpretation through doctrine and case law \cite{fritz2024notion}. These originality tests provide limited guidance for creative work that emerges through collaboration or technological mediation, particularly when human contributions are indirect or difficult to trace in the output \cite{xiao2023decoding, abbott2023disrupting}. While AI-driven systems themselves cannot be authors under copyright law, outputs generated entirely by AI are excluded from protection \cite{miami_ai_authorship}. Yet, even human involvement through prompting or post-processing may fail to meet the originality threshold when generative systems substantially shape the expressive features of the work \cite{becker2024ai}. Consequently, certain forms of human–AI creative collaboration may fall outside copyright protection altogether.

Taken together, these discussions suggest that EU-copyright law relies on a limited and often ambiguous set of indicators for identifying creative contribution. Furthermore, there remains little understanding of how artists themselves reason about the distribution of ownership in human–AI collaboration, particularly when creative work unfolds across stages such as ideation, input selection, iterative refinement, and algorithmic generation.
%

%%========================================================================
\section{Preliminary Interview Study: Creative Work with AI}

To gain a better understanding of artists' perceptions of ownership, specifically in the context of creative work with AI, we first conducted a preliminary semi-structured interview \cite{lazar2017research} with two academic artists, AA1 and AA2 (1 male, 1 female), both with 22 years of work-experience as digital artists. We asked them about their approach to creation, meaningful aspects of creative work, use of AI in creative practices, and views on ownership (detailed questions from this phase are provided in Appendix~\ref{appendix:background_questions}).
The interview recordings were transcribed, verified, and anonymized by the first author. 
For analysis, we used the emergent coding method followed by thematic analysis \cite{lazar2017research}. We used MAXQDA\footnote{https://www.maxqda.com/} to code our data.

Since "authorship" in art remains disputed \cite{cronin2012collaboration}, as discussed in Section~\ref{sec:background}, we used the term "ownership" in our study.
We were also aware that "authorship" is the term used among artists. As such, for our interview questions, we always used the phrase "authorship or ownership" whenever we asked about ownership. The purpose of this was twofold: first, to avoid making artists conform to our choice of word; and second, to invite discussion about whether authorship or ownership is the right word.

\subsection{Results}
The thematic analysis led to the identification of two key themes:

\textbf{Concept Determines Authorship}:
AA1 and AA2 argued that even for collaborative work involving other people, it is still the persons who proposed the concept who are the authors of the artwork. According to AA2, artists are always, intentionally or unintentionally, influenced by others. So even when data from other people’s work goes into the piece, that should not affect ownership: \textit{"They borrow elements and imitate styles of other artists. Feeding data from other people’s work into the machine has a similar effect. So, using training data provided by others should not pose special problem to ownership".}
AA1 further referred to a piece of interactive art they had worked on, in which viewers were invited to destroy a violin to observe how many would do so. While the viewers served as collaborators, AA1 maintained that they were not owners of the artwork because they had not contributed to the underlying concept.

\textbf{Metaphors and Comparisons:}
AA2 pointed to the history of fine arts from a different perspective: it was commonplace for artists throughout history to rely on apprentices and assistants for their work. Even Michelangelo, known for being a solitary artist, relied on apprentices while working on the Sistine Chapel. Yet it is Michelangelo, not his apprentices, who is recognized as the artist behind the work.

In summary, we realized that our interviewees did not perceive AI as introducing new challenges to ownership. This view stemmed from their belief that ownership is primarily determined by the underlying artistic concept. 
However, artistic ownership is something widely debated in academia and society as a whole. As such, we wanted to provoke our participants to engage more with the issue. In particular, we wanted to find out how they would react if they were forced to give up a part of their ownership to AI under an algorithm that quantifies artistic work.
This prompted us to explore how artists would respond to an AI-driven system that quantifies ownership.

\section{ArtSplit: A Provotype for Attributing Ownership in Human–AI Creative Collaboration} 
\begin{figure}[h]
    \centering
    \includegraphics[width=0.90\linewidth]{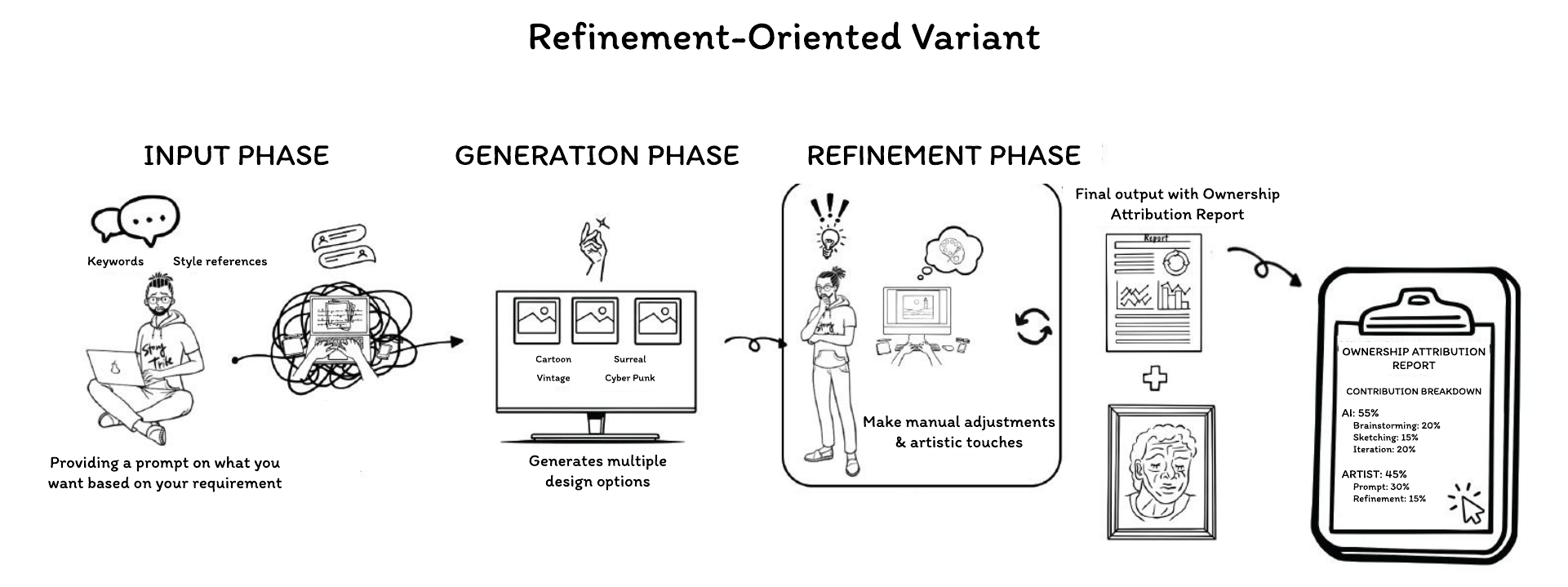}
    \caption{\textit{Refinement-Oriented Variant} co-creation process between the artist and ArtSplit, where the artist provides a prompt, receives an image, refines it, and is assigned approximately half of the ownership.}
    \label{fig:scenarios_sketch_1} 
    \Description{This figure illustrates the scenario involving the Refinement-Oriented variant. The image shows the workflow from left to right. The artist enters a prompt, receives an image, refines it, and is assigned ownership scores.}
\end{figure}

\begin{figure}[h]
    \centering
    \includegraphics[width=0.90\linewidth]{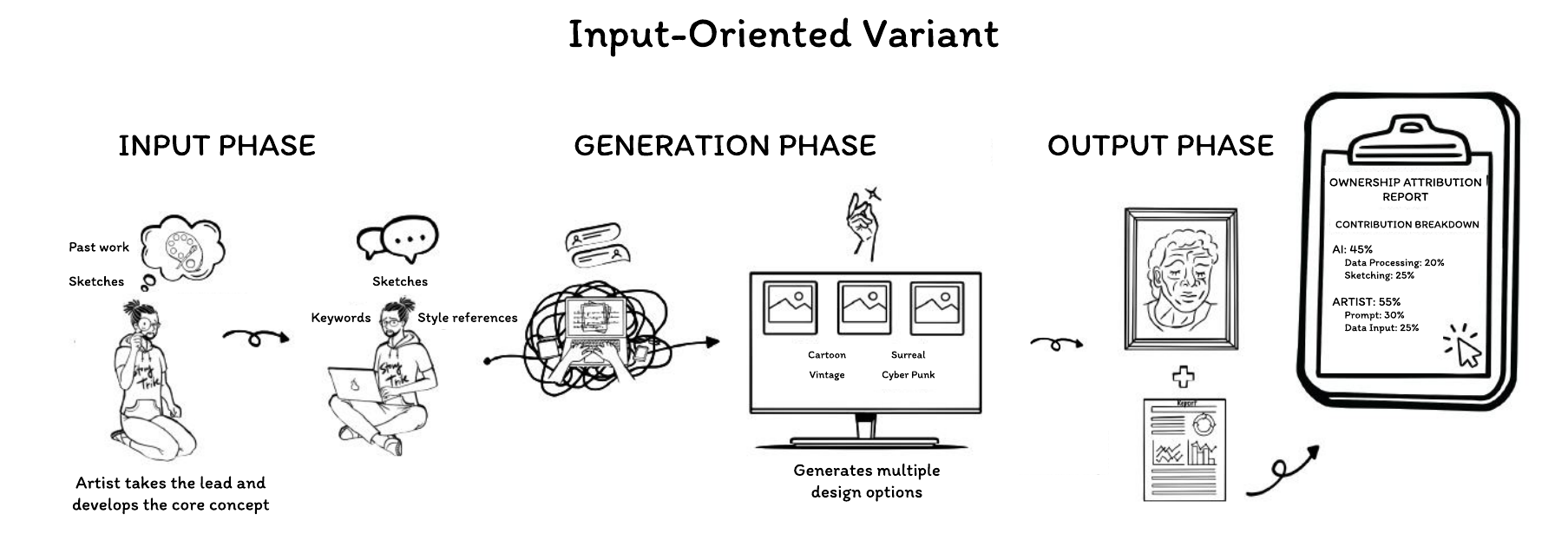}
    \caption{\textit{Input-Oriented Variant} co-creation process between the artist and ArtSplit, where the artist provides a prompt, uploads previous work, receives an image, and accepts it without further refinement.}
    \label{fig:scenarios_sketch_2} 
    \Description{This figure illustrates the scenario involving the Input-Oriented variant. The image shows the workflow from left to right. The artist provides a prompt, uploads previous work, receives an image, does not refine it, and is assigned ownership scores.}
\end{figure}

\begin{figure}[h]
    \centering
    \includegraphics[width=0.90\linewidth]{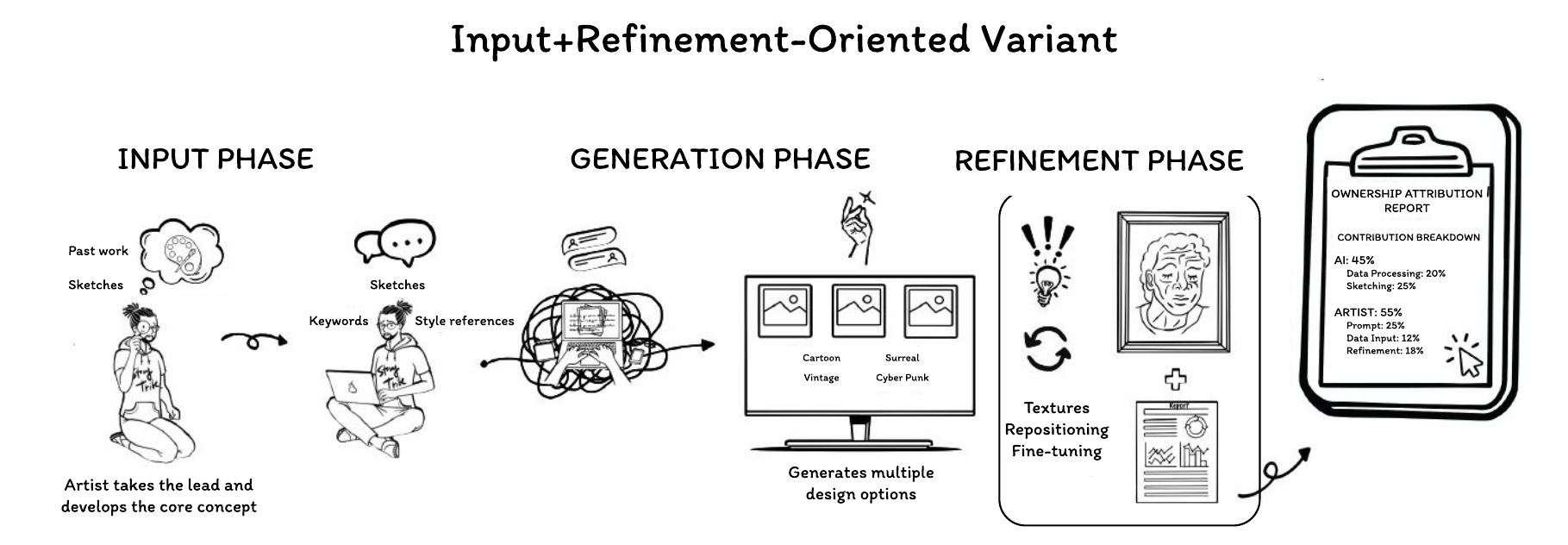}
    \caption{\textit{Input+Refinement-Oriented Variant} co-creation process between the artist and ArtSplit, in which the artist provides a prompt, uploads previous work, receives an image, and then refines it.}
    \label{fig:scenarios_sketch_3} 
    \Description{This figure illustrates the scenario involving the Input+Refinement-Oriented variant. The image shows the workflow from left to right. The artist provides a prompt, uploads previous work, receives an image, refines it, is assigned ownership scores, refines the image again, and is assigned final ownership scores.}
\end{figure}

To explore how legal assumptions align with different forms of human–AI creative contribution, we designed three provotypes of an AI-driven system named \textit{ArtSplit} that make such assumptions visible and open to critique. In three different variants, this system tracked the actions performed by the artist and an AI image-generator, and offered a breakdown of ownership in terms of the type and amount of work done by each. ArtSplit was designed as a provotype \cite{boer2012provotypes} to provoke critical thinking and tackle discussions on the notion of ownership in creative work.
We brainstormed and shortlisted seven different metrics for breaking down the artistic contribution: "Prompt" (entering a prompt), "Data Input" (uploading reference images), "Data Processing" (processing input-data by AI), "Refinement" (manually editing AI-generated output), "Brainstorming" (exploring creative directions), "Sketch" (initial visual drafts), and "Ideation" (formulating core artistic concept). 
The concept of ArtSplit and its three variants was first visualized in three sketches as shown in Figures~\ref{fig:scenarios_sketch_1}, \ref{fig:scenarios_sketch_2}, and \ref{fig:scenarios_sketch_3}, and was later developed in Figma\footnote{\url{https://www.figma.com/}}. The interaction with ArtSplit variants was enacted and presented through video vignettes. In the following, we present the variants in detail.

% ----------------------------------- Variant 1 -------------------------------------------
\textbf{Variant 1: Refinement-Oriented Variant}
In this variant, the artist entered a prompt and refined the generated image. These two actions (prompting and refining) were the basis for assigning ownership.
This variant aligns most closely with EU-copyright doctrine, which identifies authorship through free and creative choices that can be seen in the final form of a work \cite{infopaq2009, painer2011}.
The AI image-generator interpreted the prompt (labeled \textit{brainstorming}), produced an image, and responded to refinement commands (labeled \textit{iteration}). After the co-generation is completed, ownership attribution values displayed to the artist are: Artist 45\% (prompt 30\%, refinement 15\%) and AI 55\% (Brainstorming 20\%, sketch 15\%, iteration 20\%). We chose category names that felt intuitive (such as prompting, processing, and refinement) and also intended to prompt participants to critique whether these terms accurately reflected the actions shown.
Figure~\ref{fig:scenarios_sketch_1} illustrates this variant, showing how digital artists co-create art with ArtSplit.

% Scenarios Sketch #1

% ----------------------------------- Variant 2 -------------------------------------------
\textbf{Variant 2: Input-Oriented Variant}
This variant focuses on situations in which the artist contributes through prompting and by providing reference images, but does not further modify the generated result.
Ownership values after the co-generation was completed, were set to: Artist 55\% (prompt 30\%, data input 25\%), and AI: 45\% (data processing 20\%, sketch 25\%. 
This variant reflects a persistent tension in EU-copyright law: while human intellectual input and creative intention are central to authorship, the law provides limited guidance on how such contributions should be recognized when they influence the work indirectly and without visible modification of the final form, given the longstanding focus on expression rather than ideas in EU-copyright law \cite{xiao2023decoding, rosati2013originality}.
Figure~\ref{fig:scenarios_sketch_2} illustrates this variant, showing how digital artists co-create art with ArtSplit.

% Scenarios Sketch #2

% ----------------------------------- Variant 3 -------------------------------------------
\textbf{Variant 3: Input+Refinement-Oriented Variant}
This variant emphasizes how ownership attribution can change during the creative process as additional actions are taken. The artist initially contributes through a prompt and by providing reference images (labeled \textit{prompt} and \textit{data input}), while refinement is introduced at a later stage. The AI image-generator processed both (labeled \textit{data processing}), generated a sketch, and displayed with an initial assignment of ownership, which skewed significantly towards the AI: Artist: 35\% (prompt 20\%, data input 15\%), and AI: 65\% (data processing 30\%, sketch 35\%). After refinement, the artist gained an additional metric called refinement, while the AI received no new metric, as its role remained limited to processing input. The final ownership values displayed after co-generation were: artist: 55\% (prompt 25\%, data input 12\%, refinement 18\%), and AI: 45\% (data processing 20\%, sketch 25\%).
By presenting ownership as something that changes as different forms of contribution accumulate, this variant exposes a limitation of copyright frameworks, which generally assess authorship at a single point in time and offer no clear rules for revising ownership as creative roles evolve \cite{xiao2023decoding, becker2024ai}.
As before, the naming of variants was intended to provoke critical reflection from participants.
Figure~\ref{fig:scenarios_sketch_3} illustrates this variant, showing how digital artists co-create art with ArtSplit.

% Scenarios Sketch #3

\section{Exploring Artistic Ownership Through ArtSplit}
\begin{figure}[h!]
    \centering
    \includegraphics[width=1.0\linewidth]{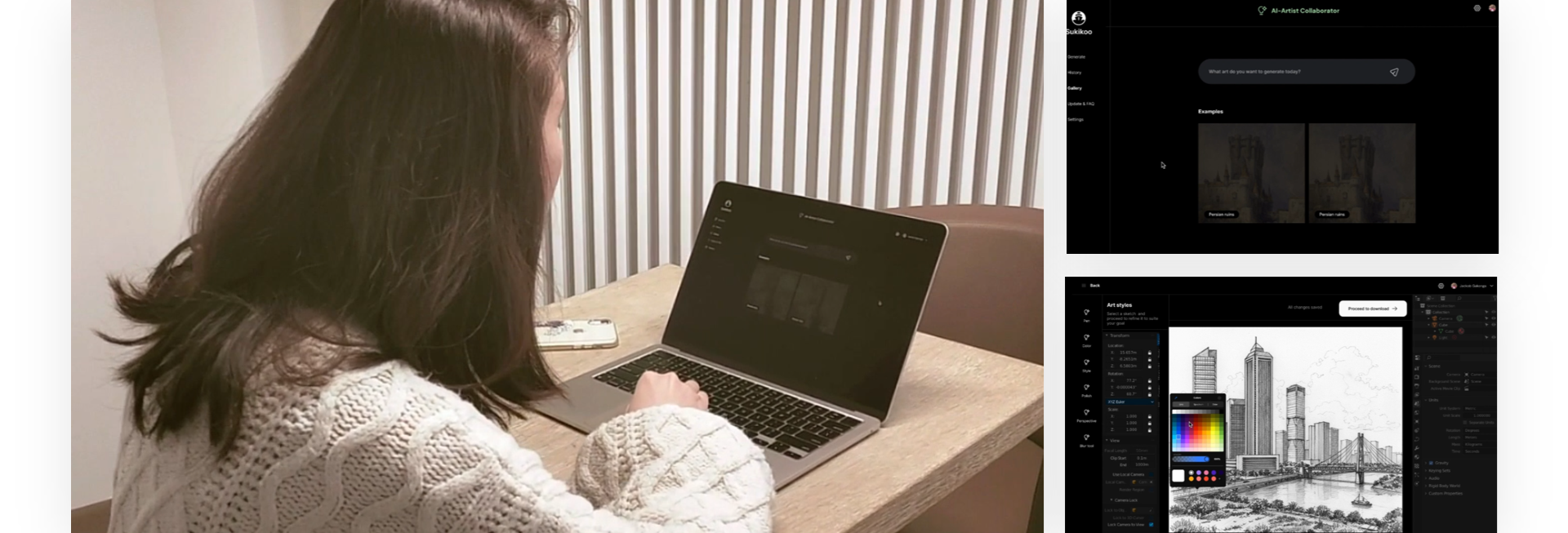}
    \caption{Left: Scenario showing an artist interacting with the ArtSplit provotype; Right (top): ArtSplit home page; Right (bottom): Editing and refining a created picture with ArtSplit.}
    \label{fig:scenario_setup} 
    \Description{This figure shows a collage of three images. The first image from the left stretches till the center and shows an artist interacting with ArtSplit provotype. The two images to its right are positioned in a single column: the top image shows the ArtSplit home page; the bottom image shows the screen in ArtSplit where the picture can be edited and refined.}
\end{figure}

We aimed to gather in-depth reflections from artists on the ownership breakdowns presented in our provotypes. To this end, we demonstrated the ArtSplit provotype through three video vignettes (Figure~\ref{fig:scenario_setup}) and reflected with participants via semi-structured interviews \cite{lazar2017research}. Each video depicted a slightly different extent of involvement of the artist in the art creation process, and the system displayed subsequent differences in ownership breakdown metrics.

% AARC Provotype & user-interaction photo

The interviews consisted of two parts. For the two academic artists who participated in the preliminary study, the goal of the first part was to recall key aspects of the initial conversation, reflect on what they previously shared, and invite additional discussion.
For the newly recruited participants, we went through the interview questions used in the preliminary study to gain a sense of their background and how they originally conceived artistic ownership.

The second part was identical for all five participants - we showed them the video vignettes featuring three different scenarios and asked them to reflect on what they observed. We sought to understand how they interpreted the differences between the scenarios, which outcomes they perceived as appropriate, and the reasoning behind their choices. We also explored how they would modify the system (if at all), what metric would be ideal for them, and how it would support their work. Participants considered the fairness of the outcomes, particularly when efforts differed between scenarios while the results remained the same, and reflected on any changes in their perception of ownership. Lastly, we asked about their willingness to use AI in their creative process if it meant sharing ownership (detailed questions from this phase are given in Appendix~\ref{appendix:reflection_questions}).

Based on minimum sample size recommendations for case studies \cite{onwuegbuzie2007typology, creswell2015educational}, we conducted semi-structured interviews \cite{lazar2017research} with five participants (2 female, 3 male, 0 diverse) aged between 26 and 53 years (M = 37.8, SD = 10.44). Three were academic artists (AAs) with 22--34 years of working experience, and two were content creators (CCs) with 4--11 years of working experience. Participants were based in Africa (n=2) and North America (n=3) and were interviewed separately via Zoom\footnote{\url{https://www.zoom.com/}}. They were recruited through the authors’ professional and industry contacts. AA1 and AA2 were interviewed twice: first as part of the preliminary study, where we asked them about their backgrounds and opinions on AI and artistic ownership; and a second time for their thoughts on interacting with ArtSplit. Each of these interviews lasted 45 minutes. AA3, CC1, and CC2 were interviewed for one hour; in the first half, we asked about their background and opinions on AI and artistic ownership, and after 30 minutes, we showed them the video vignettes. Participants were asked to fill in a consent form informing them about data anonymization and the use of recordings solely for research purposes before taking part in the study. The study was approved by the ethical board of the University of Siegen before commencement.

The interview recordings were transcribed, verified, and anonymized by the first author. For analysis, we used emergent coding method followed by thematic analysis \cite{lazar2017research}. MAXQDA\footnote{\url{https://www.maxqda.com/}} was used to support the coding process.

\subsection{Results}
During our analysis, we recognized that academic artists (AAs) and content creators (CCs) had very distinct perspectives and approaches to the provotype. Therefore, we categorized the results according to the professions of the participants. However, this does not imply that the results are representative of the professions as a whole. This is mainly an organizing scheme for this paper.
The thematic analysis of the interviews led to the identification of five key themes:

\subsubsection{\textbf{Artistic Concept Determines Authorship}}\hfill\\
From the content creators' perspectives, authorship is determined by artistic concept. Content creators applied metaphors to illustrate their point. CC1 and CC2 emphasized that AI is just a tool to speed up the process. The user decides ultimately what image to make, when it is satisfactory, and when the generative process is concluded, according to their concept. So the user is still the proper owner of the artwork. CC1 relied on comparisons to large collaborative processes such as building a company or making a film: \textit{"Say Jeff Bezos or those big corporates ... When [he] began and asked for opinions [from] friends for their companies, if their friend gave them the best insight [that] doesn't mean it's [their friends']. It was his idea!"}
However, the CCs disagreed on how exactly the artists’ concepts or ideas are conveyed in interaction with generative AIs. While CC1 believed that the artist’s idea is simply the entered prompt, CC2 argued that every act the artist performs, in addition to the prompt - all adjustments, refinements, selecting the artwork to feed the machine, and deciding when the work is complete - counts as conveying the concept. 

The academic artists held divergent perspectives on what determines ownership. While AA1 and AA2 had the same belief as the CCs (that ownership is defined by the underlying concept), AA3 was critical of the claim. AA3 questioned this paradigm using the metaphor of factory workers and owners: \textit{"Workers are the ones who have calloused hands and are there all day. But factory owner could say, I own the machines and space, and by owning the concept, I own the work. You're putting yourself in a tricky moral position."}

AA2 also questioned when assigning artistic ownership is important: \textit{"[Say] I need a frog for my article. Now, whether it's in the prompt or changing the style or doing something else, would [that] count as some sort of labor or creative input? I don't know [if] the percentage helps".} To them, when AI is used to generate something for functional purposes, it is not clear what it accomplished to calculate authorship.
AA2 suggested that the artist’s concept is something beyond the acts performed to operate the machine because "\textit{[what] the prompt does is giving an instruction that is somewhat general"}, while the concept is not identical to prompting, or any other perceivable act. More importantly, they believed that concept and instruction are distinct things: giving instructions does not always involve conveying a deeper intention or idea behind, and conveying a concept usually does not take the form of giving a series of instructions.

In summary, most participants emphasized the role of the underlying concept in determining ownership, with AA3 expressing a divergent view; while the majority attributed ownership to the concept provider, AA1 also suggested that contributors whose data was used to train the image-generator could merit a partial share of ownership (approximately 25--30\%).

\subsubsection{\textbf{Quantifiability of Ownership}}\hfill\\
After being shown the video provotype, we asked participants if they found the distributions of ownership fair. 

The content creators both thought that the artist overall should be given a greater number. CC1 said that "\textit{if the artist's own work is being fed into it, he should get at least 70\%.}" CC2 thought that refinement ought to receive more weight in all variants, while prompting and sketches were overrated. However, their tones were uncertain, positions were shifty, and explanations were based on intuition. In certain cases, they formed a sense of fairness based on the comparisons among the scenarios. For example, they found it "unfair" that the artist got a similar distribution in the Input-Oriented Variant and the Input+Refinement-Oriented Variant because the artist performed more work in the latter. Other times, their reasoning was based on reactive feelings. For example, CC1 expressed that the artist using the Input-Oriented Variant was \textit{"being lazy"}, and \textit{"taking advantage of AI, while AI does all the work"}. When confronted with their earlier claims — that concept determines ownership — the content creators made sense of their experiences in two different ways. CC1 had come to view AI more as a collaborator than a tool and believed it could, in some sense, hold a share of ownership in the artwork produced. And CC2 suggested that what the prototype tracked may not be ownership, but rather contribution, which is distinct from ownership. They noted that the prototype calculated the contribution in a “fair” way, as long as the numbers reflected the actual work done.  

All academic artists questioned what it meant for an assignment of ownership to be fair. Their responses pointed to broader concerns about the quantifiability of ownership, or contribution, in the creative process. AA1 questioned whether it was even possible to quantify the amount of work done by genuinely random processes: "\textit{Can there be fairness in something that is 100\% random, where there is an equal chance of getting just about anything?}" Using the metaphor of rolling a die, AA1 explained that \textit{"sometimes it takes multiple rolls to get the right image, and sometimes you just happen to roll the desired number right away!"} In the former case, the machine may appear to have done more work, but not because it is more productive or diligent — rather, chance is simply working against the user. It remains unclear what it would mean to fairly assess the AI’s contribution under such conditions.

About the video vignettes, AA1 pointed out that they were not given enough information about the images fed into the machine: \textit{"To assess whether ArtSplit has fairly calculated how much 'data processing' it did, we need to know what kind of images were fed into it, whether they are significantly different from the desired result. Only then could we assess whether ArtSplit has generated an image significantly different from the images fed into it or has simply picked something very similar to the output and merely done some small alterations. Knowing this is necessary for assessing how much 'data processing' it did."}

AA2 questioned whether there can be a definitive way to weigh the contribution: \textit{"What kind of justification could there be for giving prompting as 20\% of the work instead of 30\%?"} In other words, how could the work of writing the prompt be commensurable with, for example, data processing done by the machine? AA2 suggested that the weighing of the contribution could be made customizable - the user should be able to set how much the prompt would count. However, AA2 still insisted that many important elements of the creative process cannot be adequately tracked by the computer, especially when it comes to capturing the concept, as discussed in the previous theme.

AA3 pushed the question further when we asked them which steps in the creative process authorship lies: \textit{"You must assume that there are steps in the creative process. And you must assume that authorship sits in specific steps rather than sort of an overall presence in it."} AA3 explained that the 'concept' never ceases to develop until the artwork is completed; each new move could lead the artist to review their concept. To them, the artistic process is more like a multi-nodal feedback loop than something with separate predetermined steps. 

Overall, the content creators tended to accept ownership as quantifiable and favored assigning greater ownership to the artist, often reasoning through intuition and comparisons across scenarios, whereas the academic artists remained deeply skeptical of quantification, grounding their critiques in both the technical knowledge of AI image-generation and understandings of the creative process.

\subsubsection{\textbf{Appropriation}}\hfill\\
Whether or not participants believed that ArtSplit tracked ownership as intended, they envisioned ways it could be helpful. For content creators, ArtSplit could serve motivational purposes, even when they suggested that it should be seen as a tracker for contribution instead of ownership. CC1 said that it would motivate them to not be 'lazy', simply entering prompt after prompt and letting the machine do the work. Knowing that the machine has a measurable contribution to the product, CC1 thought it had become more important to include something that would make the work their own. For example, they mentioned that they would make sure to include their artistic signature in the final product. CC2 also mentioned that ArtSplit could make them more aware of the creative process and the different kinds of work that went into it, be it from the human or the machine. It would encourage them to reflect more deeply on the implications of using AI in art.

Academic artists tended to be more deliberate in interacting with ArtSplit. AA1 entertained the idea of using it as a way of providing a degree of narrative or critique to it. \textit{"How could I actually nudge the numbers to go the other way to the point where the ownership - the personal human ownership - is minimal? We've taken this idea and we've abstracted it entirely to as far as the machine can go while still having to be about the concept, the idea, or the critique!"}
AA2 thought that ArtSplit would be useful for tracking differences that might help avoid copyright conflicts. For example, if the use of an image does not violate copyright if it is altered by more than 40\%, then ArtSplit could be used to track that difference. Also, in the case of designing something solely to have it stand out, such as a cover for a book, it can again be used to track how different a particular design is compared to the pre-existing ones. They also commented that in such cases, authorship no longer mattered: \textit{"It does the purpose, but I don't think there's any desire necessarily [for] authorship ... I think it's more [of] a desire to fulfill a function."}

Overall, the CCs believed that ArtSplit would be useful for motivation and comprehension purposes, indicating possible trust in the system's potential for providing meaningful assessment of ownership. On the contrary, the AAs were more skeptical of ArtSplit's ability to fulfill its intended function and came up with their own ways of using it.

\subsubsection{\textbf{Controllability}}\hfill\\
Controllability emerged in the discussions in two ways: first, whether ArtSplit’s current design could offer artists a sense of control during the creative process; and second, whether ArtSplit itself could be designed to afford greater control.
Among the content creators, CC2 mentioned that they would like to use ArtSplit in their own work because without it, they felt a lack of control: "\textit{I don't want to feel like [...] the image-generation tool is beyond me."} It offered them a glimpse into what was happening inside the blackbox of image-generation. They believed it would give them a sense of confidence and stability during the creative process. 

The academic artists were more skeptical about the controllability provided by ArtSplit. AA3 questioned whether ArtSplit might give users a false sense of control: while the breakdown of contributions allows users to see what they have done, the categories measured are limited to those that ArtSplit is able to track. Over time, the creative actions that artists are motivated to take could become limited to those measurable by the system, unintentionally narrowing the range of methods used in the creative process. AA3 also suggested that the breakdown should be made visible to the user throughout the process, like how a word-counter can visibly track the number of words written as they are being typed. That would give the artist real-time awareness of what work has been done and how much.

\subsubsection{\textbf{Perverse Intention \& Wicked Case}}\hfill\\
There are two situations in which the purpose of ArtSplit is subverted. First, the numbers themselves become the incentive, and second, the numbers lose meaning.

We created ArtSplit under the assumption that artists would like to have their sense of authorship (or ownership) affirmed. We also assumed that authorship (or ownership) is proportional to the acts performed during the creative process (inspired by \cite{xu2024makes, hwang202580}). The consequence of designing under such an assumption is that the numbers could be manipulated by simply performing more actions, even when these actions are not desirable in themselves. The creative process is derailed and ceases to be genuinely artistic when the artist is incentivized to perform actions just to get a higher number for their contribution.

The opposite scenario, where the numbers are no longer meaningful to the artist, occurs when the artist obtains an image that meets their expectation exactly. Given that there is always a degree of randomness in the process of AI image-generation, this could happen very early on, when the artist has done very little work and hence made relatively little contribution. When asked whether they would continue to work on the image simply to get a higher number, all said that they would not, and that they would not care about the number: \textit{"It's not about me getting 55\% credit but getting the artwork I wanted"} (CC1). 

This suggests that all participants - academics and content creators - had an underlying sense of a different paradigm of ownership, where authorship is not determined by the amount of work done, or the concept conveyed, but instead by how much the author can see their own vision in the resulting image.

\section{Discussion}
In this critique paper, we aimed to explore the implications of communicating ownership breakdown in human–AI creative collaboration, with a focus on how artists perceive and understand ownership. Based on two preliminary interviews, we developed a provotype called ArtSplit that communicated ownership over a co-generated image via three variants: refinement-oriented, input-oriented, and input+refinement-oriented. We then conducted semi-structured interviews with five artists to observe their reactions to these different breakdowns and gain insight into their understandings of ownership (or authorship) and the factors that shape their perceptions of it. In the following, we discuss our results.

\subsection{Contribution v.s. Authorship}

An implicit assumption that went into the design of ArtSplit was that authorship is distributed according to contribution. However, our participants considered only certain types of contributions to be relevant to authorship, particularly those involving the communication of concepts and the work required to ensure that the results aligned with the artist's intent.
The basis for this perception likely rests on the fact that an artwork is at once a piece of art and an object. Although different kinds of work are necessary to bring the artwork into being as an object, not all contribute to its artistic value. Hence, what ArtSplit actually tracks was not perceived as authorship but as contribution.

Collaborative artistic practices have existed since the Renaissance \cite{workbernardino, horne1995craft}. Artists today still employ assistants to do the grunt work in artistic production and setting up complex installations. The \textit{Auteur} Theory in Film Studies poses the director as the main creator, or the \textit{auteur} (French word for author) of the production, although filmmaking tends to be intensely collaborative \cite{Responding2002}. Furthermore, movements within modern art alongside the technologies enabling this (especially AR and VR) gave rise to interactive art: a kind of artwork that only becomes actualized with viewers' participation. In such cases, viewers themselves become collaborators, though often they are ignorant of the artist's intention or what to expect. In these examples, we do not tend to dispute who the author is. Thus, to claim that collaboration with AI causes special problems for ownership, we must articulate how exactly AI differs from other human collaborators.

\subsection{Tracking Artistic Processes}
The participants' diverging understandings of “artistic concept” prompted us to question whether tracking artistic acts is necessary or even feasible in creative practice.
First, if concept determines ownership, then, within this understanding, any work that does not qualify as an act of “conveying a concept” may not be considered relevant or counted towards ownership. 
Second, even if we interpret the breakdown as tracking contribution rather than ownership, questions remain regarding which acts should fall under the category of “conveying an artistic concept” - whether this includes only prompting or all actions performed by the artist. 
Third, if none of the acts tracked by ArtSplit qualify as conveying a concept — as suggested by AA2, who compared prompting to giving instructions rather than expressing concepts — then an indispensable component may be omitted by the system.
This suggests that AI-driven systems may face challenges in capturing certain aspects of the creative process.

Another question we should pose is what it means to quantify the components of creative work, given that the categories participants discussed appeared deeply entangled in ways that may be difficult to capture with computational artifacts.
The concept may change continually in the background as the artist engages in other parts of the work. However, systems such as ArtSplit could track refinement, but not what is happening in the artist’s mind. 
A further question arises when we consider the fact that one's prompting skill gets better as one becomes more experienced in writing prompts. 
Consequently, an experienced artist may achieve a desired outcome with fewer prompting iterations. Yet, within participants' understandings of contribution, fewer observable acts would not necessarily imply reduced contribution or authorship. This resonates with the contemporary adage: \textit{you are not paying for a few minutes of work, but for the years it took to learn how to do it in a few minutes}\footnote{\url{https://medium.com/@belikenikola/the-phrase-you-dont-pay-me-for-the-5-minutes-i-spent-to-do-it-you-have-to-pay-me-for-10-years-i-f7b3064d7284}}.
One way to address these problems would be to give artists the freedom to decide which acts should weigh more, according to their own understanding of what determines authorship. However, this would introduce further challenges. On one hand, when artists are not allowed to set their own metric, the system may yield results that conflict with their understanding. On the other hand, if they are free to set the metric however they like, then the numbers could become meaningless, as artists might simply configure the system in a way that always designates them as the primary author.

\subsection{When Originality Matters}
Another presupposition we relied on for designing the provotype was that originality matters: it is because the artist may want to know how much of their creativity has gone into the resulting artwork that they may take an interest in using an AI-driven system like ours. However, depending on the purpose of the artwork, authorship matters less in some cases. As AA2 pointed out, authorship is usually not desired when the piece serves to fulfill certain functions, such as meeting marketing demands or avoiding copyright conflicts. This issue could be seen as another consequence of conflating ownership with contribution: in some contexts, users may primarily seek acknowledgment of their contribution, without needing to determine whether anyone should be considered the author of the piece. Insisting on deciding who the author is in every case can therefore give rise to questions that are difficult to answer because they need not be asked in the first place.

Furthermore, tracking contribution does not imply tracking originality: the fact that work is being done does not say anything about its originality. AA1 suggested that if ArtSplit were able to compare the resulting image with all images on the internet, then it might have been capable of indicating the originality of the artwork. However, we suspect that even that would not be enough, because originality may not mean mere differences. Consider Duchamp's L.H.O.O.Q., a postcard of the Mona Lisa with facial hair\footnote{\url{https://www.nortonsimon.org/art/detail/P.1969.094}}. The image looks almost exactly like the Mona Lisa, but the little beard uniquely reflects Duchamp's artistic character.
From this perspective, creative work, being a kind of conscious activity, is not novel because it happens to be unlike anything previously created, but rather because of the unique artistic concept of the creator \cite{stein1953creativity}.

\subsection{Making Sense of The Numbers}
A good design, according to HCI guidelines, affords transparency and explainability, usually through enabling intuitive comparisons \cite{guo2024exploringimpact, oladele2024generative, xu2024makes}. However, we question whether systems like ArtSplit offer merely an illusion of transparency. 
In our case study, participants' levels of AI literacy appeared to shape how they criticized the numbers generated by ArtSplit.
The Academic Artists were aware that the numbers and categories being measured were made-up, and used that as the starting point for their critique. They discussed the possibilities of quantification and how the numbers could be made compelling or useful. Meanwhile, the CCs assumed we knew what we were doing and hence believed that the numbers were meaningful and contributions were quantifiable. The CCs did not have a basis for questioning the feasibility of ArtSplit, but could only express dismay about the specific assignments of contribution in certain scenarios. In other words, the academics were able to question whether a "fair" assignment of contribution and authorship was even possible, whereas the CCs assumed that it was possible and hoped that the numbers could be nudged in a direction that affirmed their intuition.

We intended to create a sense of ownership by providing a numerical breakdown of contributions, and through that, offer the user a more transparent experience of artistic creation. But this sense of transparency could be an illusion--it is when the user is capable of knowing, independent of such a system, how each category of contribution ought to be measured, that the numbers could be truly meaningful and convincing to them (in the same way that, for someone who does not understand arithmetic operations, the outputs of a calculator are mysterious).

The observations noted in this section also give rise to an ethical issue: Is it permissible to test prototypes that appear to measure things that might not be measurable with users? Before taking part in our study, the CCs had never thought about whether artistic contribution could be measurable. But after the interviews, they seemed to believe that it is possible to do so, just because we showed them such a prototype while posing as scientists. The interviews may have given them false beliefs.

\subsection{Design Less?}
The final point we arrive at is: Is an AI-driven system such as ArtSplit actually needed? Does it address an actual problem, or merely the symptoms of deeper structural issues? Literature critiquing technosolutionism has noted that the implication of case studies is "not to design (technology)" \cite{2017notdesign}. Designing when technology is not the best solution could transform the issue in unanticipated ways, or lead researchers to solve \textit{"a computationally trackable transformation of a problem rather than the problem itself"} \cite{2011notdesign}.

In our case, it is also worth noting that the contemporary need to formally "own" the artwork is shaped by our current social-historical context.
The modern notion of copyright came to be after the invention of mass printing technologies to enable reproducible works to be bought and sold \cite{plagiarizethispaper}. Before that, texts and images were simply not considered things that could be "owned". The subjective need for feeling that one is the author or owner of something is still different from copyright. Fyre \cite{plagiarizethispaper} argues that the need among artists and academics to have their creations or ideas attributed to them is a result of power being distributed according to the production of artwork and publications. 
This raises the question of whether introducing quantification into this complex web of problems is an appropriate response.

\subsection{Concluding Remarks}
This critique paper was inspired by ongoing debates within artist communities and legislative bodies concerning authorship and ownership in AI-mediated creative work. By confronting academic artists and content creators with a provotype that explicitly quantified human and AI contributions, we sought to elicit reflective discussion and to better understand how creative workers themselves perceive artistic ownership in such collaborations. 
There are two primary limitations to our study. The first concerns the number of participants, which is smaller than what is expected in HCI research.
Our analysis was informed by Interpretative Phenomenological Analysis (IPA), a methodology that values the subjective and situated experiences of individuals \cite{smith2021interpretative}. Accordingly, we prioritized rich, nuanced accounts over generalizability. In line with other qualitative approaches \cite{yin2011applications}, our concern was with the existence of specific experiences rather than their frequency. The small sample size is further justified by the depth of engagement required and the logistical complexity inherent in conducting such qualitative research.

The provotype used in this case study was presented through video vignettes \cite{matza2021vignette}, which, while useful for provoking reflection, may not have fully captured the complexity of real-world creative practices. The vignettes were based on a generalized model of the artistic workflow, which did not account for the diversity of individual work habits. Consequently, key contextual factors may have been overlooked, and participants’ immediate reactions—especially during time-limited interviews—may not reflect their deeper or evolving perspectives. Nevertheless, despite the aforementioned limitations, this methodology enabled the possibility to confront participants with a not-yet-existing practice.

Taken together, this work does not so much resolve questions of ownership in AI-assisted creativity as it unsettles the assumption that such questions should be addressed through system design in the first place. By making ownership explicit (and measurable), systems like ArtSplit surface tensions that extend beyond interface or metric choice, pointing instead to historically contingent assumptions about authorship, value, and creative labor. In this sense, our provotype does not argue for better ways of quantifying ownership, but invites reflection on whether quantification risks reifying and operationalizing concerns that are themselves products of broader social, legal, and institutional arrangements. We therefore position this work not as a call to refine ownership-tracking systems, but as a critique of the impulse to computationally resolve questions that may resist technical solution altogether.

%%
%% The acknowledgments section is defined using the "acks" environment
%% (and NOT an unnumbered section). This ensures the proper
%% identification of the section in the article metadata, and the
%% consistent spelling of the heading.
% \begin{acks}
% \textcolor{red}{We thank the participants and colleagues whose support and valuable input contributed to this work.}
% \end{acks}

%%
%% The next two lines define the bibliography style to be used, and
%% the bibliography file.
\bibliographystyle{ACM-Reference-Format}
\bibliography{sample-base}

%%
%% If your work has an appendix, this is the place to put it.
\appendix

\section{Appendix: Interview Questions}

\subsection{Preliminary Background Questions}
\label{appendix:background_questions}

\begin{enumerate}
    \item What art do you make?
    \item What makes your work meaningful to you? Why?
    \item Do you work with AI, and how has your experience been? 
    \begin{enumerate}
        \item How long have you been doing this?
        \item How do you use AI (in which steps in the creative process is AI utilized and how)?
        \item What made you decide to try AI out?
        \item What is your general attitude or experience about using AI in art?
    \end{enumerate}    
    \item How do you view AI (as a collaborator like a colleague/ a tool like a brush or software/ something else entirely)? How does that influence your view of ownership?
    \item According to you, which steps in the creative process contribute the most to ownership?
    \item Who do you think owns the final artwork (yourself as the creator/ the AI/ the AI’s developers/ shared ownership model)? Why?
\end{enumerate}

\subsection{Provotype Reflection Questions}
\label{appendix:reflection_questions}

\begin{enumerate}
    \item Reflect on what happened in the scenarios. 
    \begin{enumerate}
        \item Did you notice the differences across the three scenarios?
        \item Which result was the most appropriate for you? Why?
        \item Which variant of the system would you use? Why?
        \item What is your general attitude or experience about using AI in art?
    \end{enumerate}    
    \item Reflect on the AI-Artist Collaborator prototype. 
    \begin{enumerate}
        \item Would you change anything about the system? Why?
        \item How would you change it?
        \item What would be the best model for you? Why?
    \end{enumerate}    
    \item Reflect on a workday with such a system.
    \begin{enumerate}
        \item What do you think such a system does for artists?
        \item Would you like such a system to be introduced in your work? Why?
        \item How could it help you? 
    \end{enumerate}    
    \item You noticed that in scenario 2 and scenario 3, the artist does more work in one of them but not the other. However the contribution breakdown displayed by the system remained the same. Do you think it is correct? Incorrect? Or do you feel it does not matter? Why?
    \item Has any of your initial views on ownership of AI art changed? Why or why not?
    \item How inclined are you to use AI, now that you have to give away part of the ownership?
\end{enumerate}

\end{document}